\documentclass[letterpaper, 10 pt, conference]{ieeeconf}  % Comment this line out if you need a4paper

\IEEEoverridecommandlockouts                              % This command is only needed if 
\usepackage[linesnumbered,ruled,vlined]{algorithm2e}
\usepackage{float} % More control over graphic placement
\usepackage{amsmath}
\usepackage{amssymb}
\usepackage[english]{babel}
\usepackage[hyphens]{url}  %Required
\usepackage{blindtext}
\usepackage{fancyhdr}
\usepackage{graphicx}
\usepackage[export]{adjustbox}
\usepackage{subcaption}
\usepackage{multicol}
\usepackage{comment}

\title{\LARGE \bf
A Deep Learning Approach To Multi-Context Socially-Aware Navigation
}
\author{Santosh Balajee Banisetty$^{1}$, Vineeth Rajamohan$^{2}$, Fausto Vega$^{3}$, and David Feil-Seifer$^{4}$% <-this % stops a space
\thanks{$^{1}$Santosh Balajee Banisetty, $^{2}$Vineeth Rajamohan, and $^{4}$David Feil-Seifer are with the Department of Computer Science and Engineering, University of Nevada, Reno,
        1664 N. Virginia Street, Reno, NV 89557-0171, USA
        {\tt\small santoshbanisetty@nevada.unr.edu}, {\tt\small vrajamohan@nevada.unr.edu}, and {\tt\small dave@cse.unr.edu}}%
\thanks{$^{3}$Fausto Vega is with the Department of Mechanical Engineering, University of Nevada, Las Vegas,
        4505 S. Maryland Pkwy, Las Vegas, NV 89154, USA
        {\tt\small vegaf1@unlv.nevada.edu}}%
}

\begin{document}

\maketitle
\thispagestyle{empty}
\pagestyle{empty}

%%%%%%%%%%%%%%%%%%%%%%%%%%%%%%%%%%%%%%%%%%%%%%%%%%%%%%%%%%%%%%%%%%%%%%%%%%%%%%%%
\begin{abstract}
%Understanding the interaction context and using context-based social norms helps a robot exhibit socially-aware behaviors in multiple contexts. Robots' abilities to exhibit social behaviors in single or few contexts limit their use in real-world environments where these robots encounter not one but multiple social challenges related to navigation. In this paper, 
We present a context classification pipeline to allow a robot to change its navigation strategy based on the observed social scenario. Socially-Aware Navigation considers social behavior in order to improve navigation around people. Most of the existing research uses different techniques to incorporate social norms into robot path planning for a single context. Methods that work for hallway behavior might not work for approaching people, and so on.  We developed a high-level decision-making subsystem, a model-based context classifier, and a multi-objective optimization-based local planner to achieve socially-aware trajectories for autonomously sensed contexts. Using a context classification system, the robot can select social objectives that are later used by Pareto Concavity Elimination Transformation (PaCcET) based local planner to generate safe, comfortable and socially-appropriate trajectories for its environment. This was tested and validated in multiple environments on a Pioneer mobile robot platform; results show that the robot was able to select and account for social objectives related to navigation autonomously.
\end{abstract}
%%%%%%%%%%%%%%%%%%%%%%%%%%%%%%%%%%%%%%%%%%%%%%%%%%%%%%%%%%%%%%%%%%%%%%%%%%%%%%%%

\section{Introduction}

Human-human interpersonal navigation behavior is governed by social rules, which depend heavily on the environmental context. Robots must follow social rules governing the use of space when in close proximity to humans. A socially-aware navigation (SAN) planner could allow a robot to consider social information in order to plan its movement. Advancements in planning, control, etc. allow robots to extend their operation from a controlled lab environment to real-world dynamic environments. Changes in environment mean that the social rules governing navigation interaction might also change. For social robots to be deployed and be successful in human environments, they should be able to adapt to various interaction situations. Context-aware social behavior related to navigation is important for a successful human-robot interaction (HRI). A comprehensive solution is required for socially-aware navigation, challenges common to SAN should be dealt with holistically~\cite{kruse2013human, banisetty2018towards}. %%%

%The importance of social awareness for improving human comfort in the presence of robots has been demonstrated by several studies \cite{lasota2015analyzing},\cite{shiomi2014towards}.

% \begin{figure}[t]
% \centering
% \includegraphics[width =\columnwidth, height = 5cm]{usan_iros.png}
% \caption{An overview of the proposed Unified Socially-Aware Navigation (USAN). Modules with in the dotted lines are the modification to ROS navigation stack that we propose, blocks in blue are from ROS navigation framework. Module in cyan, PaCcET local planner is from our prior work~\cite{forer2018socially}. Blocks in green deal with context classification and objective selection are the contributions of this paper. The intent recognition module, magenta block is an on-going work.}
% \label{fig:usan}
% \end{figure}

\begin{figure}[t]
\centering
\includegraphics[height = 4cm]{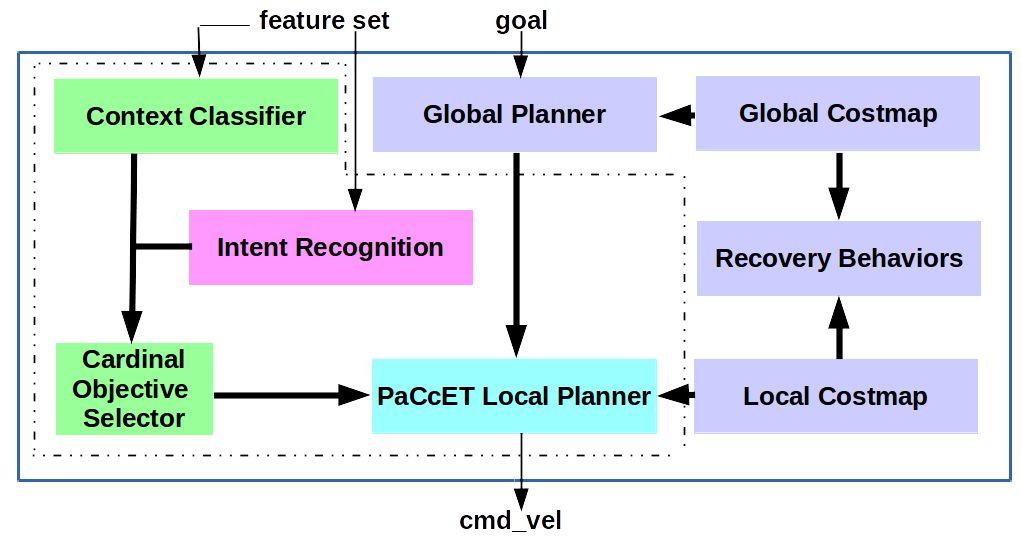}
\includegraphics[height = 3cm]{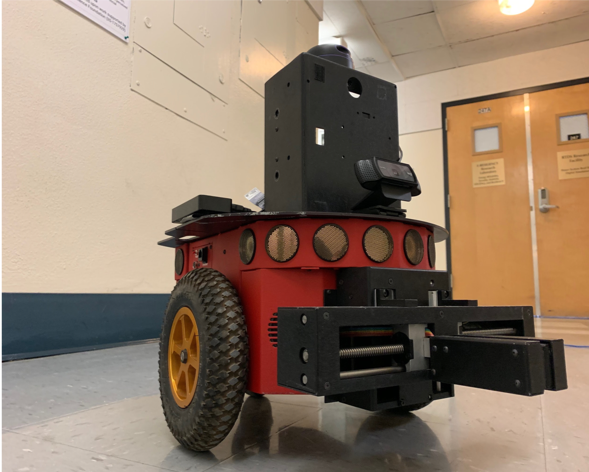}
\caption{\textbf{Left: } The proposed Unified Socially-Aware Navigation (USAN). The dotted lines are the modification to ROS navigation stack that we propose, blocks in blue are from ROS navigation framework. The PaCcET local planner (cyan) is from our prior work~\cite{forer2018socially}. Context classification and objective selection (green) are the contributions of this paper. The intent recognition module, magenta block is an on-going work. \textbf{Right: } A Pioneer robot used to implement and validate the proposed method\vspace{-3mm}}
\label{fig:usan}
\label{fig:pioneer}

\end{figure}

Our prior work has demonstrated in simulation that a local planner utilizing Pareto Concavity Elimination Transformation (PaCcET) could generate SAN trajectories accounting for personal space in a hallway~\cite{forer2018socially}. We extended the PaCcET local planner's multi-objective optimization capabilities to different contexts like art gallery interaction, O-formations and standing in a line and validated it on a mobile robot platform~\cite{banisetty2019sociallyaware}.  
We propose a learning approach using CNN and SVM to detect the on-going interaction context; we integrated the context classifier with a non-linear optimization-based local planner~\cite{banisetty2019sociallyaware,forer2018socially} to achieve context-appropriate robot trajectories. Using autonomous context classification and a PaCcET-enabled local planner, we can achieve socially-aware navigation behaviors not just for a single context but for multiple contexts. We realize and validate our unified socially-aware navigation (USAN) architecture~\cite{banisetty2018towards}. The remainder of this paper is structured as follows. In the next section, we review related works. In Section~\ref{sec:tech}, we discuss the technical details of the architecture. In Section~\ref{sec:results}, we apply our method to various scenarios on a real robot to validate the proposed approach. Finally, in Section~\ref{sec:discussion}, discussion and future directions are presented.

\section{Related Work}
\label{sec:rel_work}
%AAAI'16, RO-MAN'17

Methods for generating a collision-free path for robot navigation~\cite{marder2010office} do not include social norms in their algorithms. Incorporating social norms and proxemics into robot path planning algorithms, SAN, can help address HRI missteps~\cite{mutlu2008robots}. This is especially important in dynamic human environments. One of the early work in SAN, the Social Force Model (SFM)~\cite{helbing1995social} uses social ``forces" to consider pedestrians near a robot as external-projecting forces. This model can be extended to a group rather than just an individual to detect abnormal behavior in a crowd~\cite{mehran2009abnormal} by using a bag of words approach to classify frames as normal and abnormal. Many SAN solutions work for a single social scenario. For example, a method for hallway behavior~\cite{sebastian2017socially-aware, zanlungo2012microscopic}, a method for approaching people~\cite{satake2009approach}, a method for waiting in a queue~\cite{nakauchi2002social}. Such methods solve individual challenges, but their functionalities are context-specific. 

Time-dependent planning~\cite{kollmitz2015time} combined with layered social costmap~\cite{lu2014layered} generates plans that closely resembles a human-based interaction approaches. This method was applied to increase the efficiency of human-robot collaborative assembly tasks in intra-factory logistics scenarios by modeling assembly stations and operators as cost functions in a layered cost map. The preliminary experiment results showed that the system is capable of modeling both workspaces and operators in different layers and combine them with obstacle information~\cite{santana2018human}. The layered costmaps approach to SAN utilizes different costmaps for various contexts to perform socially-aware navigation by computing a master costmap~\cite{lu2014layered}. However, the layered costmaps approach does not include a mechanism to autonomously select the layers (costmaps) for a sensed interaction context; thus it effectively is a single context SAN like most of the related work.

%unrelated
%Approaches to SAN range from simple cost functions to deep learning based techniques involving social norms. 
%unrelated
%Most of them deal with a single interaction context, i.e., hallway navigation or approaching a person, etc.

%Similarly, the Social Force Model (SFM)~\cite{helbing1995social} used social ``forces" to consider pedestrians near a robot as external-projecting forces. That this model can be extended to a group rather than just an individual to detect abnormal behavior in a crowd~\cite{mehran2009abnormal} by using a bag of words approach to classify frames as normal and abnormal. 

%Sample-based motion planning can be useful for high-dimensional spaces, but make computation difficult. Burns \textit{et al.}~\cite{burns2005sampling} showed a significant decrease in estimation cost on motion planning by exploring the information gathered from sampling and representing that information in a predictive model. Kingston \textit{et al.} \cite{kingston2018sampling} claim that in a sample-based motion planning approach, it is difficult to include task constraints. In addition, extending sample-based algorithms to include geometric constraints in motion planning have heavily been discussed.  

Deep reinforcement learning has been used for motion planning that accounts for social norms when navigating~\cite{chen2017socially}. The robot observed and learned a policy continuously for an optimal path that will avoid collisions with humans and objects. Similar to deep reinforcement learning, inverse reinforcement learning (IRL) can plan socially-aware paths for robots based on human demonstration. By combining a feature extraction module, IRL module, and a path planning module to generate a human-like path~\cite{kim2016socially}. This method was further extended for robots to navigate in a crowded environment~\cite{vasquez2014inverse} by evaluating two different IRL approaches and many feature sets in wide-scale simulation. Voronoi graph-based IRL methods can be used to efficiently explore the space of trajectories from the robot’s start to end position~\cite{kretzschmar2016socially} for navigation in an office environment in the presence of humans. A graph-based method was applied to learn motion behavior using Bayesian IRL using sampled data~\cite{okal2016learning} shows that a robot was able to learn complex navigation behaviors. Deep reinforcement learning and IRL methods for path planning problems need a considerable amount of data, computational time, and memory for a single context, let alone generalize to multiple contexts.

Our prior work modeled human navigation behavior using a Gaussian Mixture Models (GMM) using autonomously-detected features to differentiate between various interaction scenarios~\cite{banisetty2016socially} and then extended the GMM approach to a SAN planner~\cite{sebastian2017socially-aware}. %While a model-based approach worked for local planning, it required a trained model for every interaction the robot might encounter. 
%We then developed a SAN planner that utilized the GMM based social model to generate socially-appropriate trajectories in following, passing, and meeting scenarios of a hallway interaction~\cite{sebastian2017socially-aware}. 
Taking into account interpersonal distance generates not only safe, but also comfortable social trajectories~\cite{forer2018socially}. A model-based approach works well for high-level decisions, including: what context is this interaction? What objectives are essential in a sensed context, etc. On the other hand, the optimization approach requires less computational time and is suitable for low-level local planning tasks. In the next section, we will see how a combination of a model-based decision-maker and a multi-objective optimization-based local planner can be used to achieve objectives of a unified socially-aware navigation. 

%\end{comment} 

%%% Dave read to here %%%

\section{Approach}
\label{sec:tech}
%Overview

% \begin{figure}[t]
% \centering
% \includegraphics[width =0.9\columnwidth, height = 5.5cm]{pioneer.png}
% \caption{An upgraded Pioneer robot used to implement and validate the proposed method.}
% \label{fig:pioneer}
% \end{figure}

The realization of a USAN architecture presented in this paper requires visual classification of context and laser-based detection of group configurations to select appropriate navigation behavior. The USAN architecture shown in Figure~\ref{fig:usan} is implemented and tested on a pioneer mobile robot (shown in figure~\ref{fig:pioneer}) with an upgraded camera and a long range laser setup. Appropriate behavior related to navigation can be achieved by a local planner that accounts for social normality. In this section, we discuss all the significant components of USAN architecture that include a CNN based visual context classifier, a laser-based group formation detection using SVM, and a modified local planner that utilizes non-linear optimization to generate local trajectories that are socially appropriate for an autonomously sensed context. 

%\subsection{Convolutional Neural Network}
%Before continuing with the model architecture and the dataset, we briefly describe general information on CNN, a type of neural network that is used for context classification in this paper. 

%CNN uses backpropagation~\cite{hecht1992theory} to learn the weights of the neural network that can predict different kinds of recommendations. Convolution layers, pooling layers, and fully connected layers as the core components of a CNN network architecture. State-of-the-art classification tasks~\cite{lecun1990handwritten} and end to end systems~\cite{bojarski2016end} widely used CNNs at their core. CNN use cases extend to various domains like perception, speech recognition, and autonomous vehicles, to name a few. 

%Convolution layer is at the core of a CNN, which utilizes convolution operation to extract features in an image. Pooling layers, a down-sampling process, reduces the size of the input array. There are two types of pooling mechanisms, namely, max pooling~\cite{scherer2010evaluation} and mean pooling. In max pooling technique, the maximum value in the region is picked. Whereas, in mean pooling, the average of all the elements in the region is picked. At the end of the CNN are the fully connected layers that have full connections from the previous layer to make a classification decision based on activation. In addition to the mentioned layers, CNN often uses different regularization techniques to eliminate over-fitting in its learning process; one such technique is Dropout~\cite{srivastava2014dropout}. 

\subsection{Context Dataset}
\label{dataset}
\begin{figure}[ht]
    \centering
    \includegraphics[width=0.95\columnwidth]{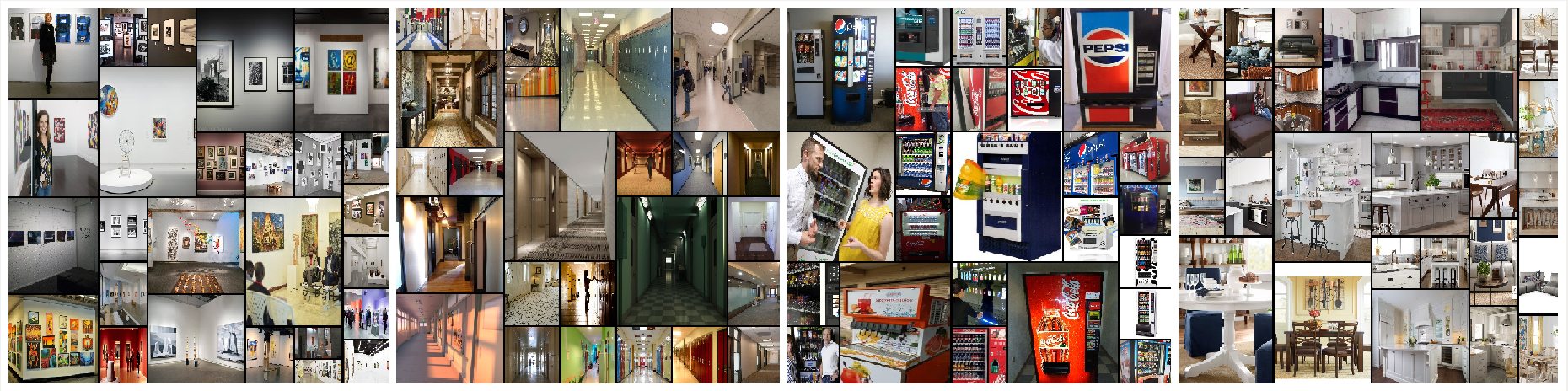}
    \caption{A sample of images from the internet that constitute images of hallways, artwork, vending machines and other categories used for training our model.}
    \label{fig:dataset}
\end{figure}

We trained a CNN model to distinguish between four contexts (classes), art gallery, hallway, vending machine and others (anything which is not a hallway, art gallery or vending machine - we utilized images of kitchens, living rooms, and dining rooms). We collected a total of 4773 images from the internet as shown in Figure~\ref{fig:dataset} and split them into training (.75), validation data (.25) and further kept aside 400 images for testing on the model as shown in Table~\ref{table:dataset}. The images collected were all in color, resized to 256x256 and normalized before feeding to the network. As the dataset is relatively small, data augmentation was incorporated to ensure model generalization. Augmented data includes image manipulations like zoom, shear, a shift in width, a shift in height, horizontal, and vertical flip. 

Apart from the data collected from the internet, we collected real-world data at the University of Nevada, Reno to further test the model. The locations on campus, where we collected data, include buildings in the Colleges of Engineering, Science, and Humanities. The real-world data used for testing, but not part of the training process includes all the classes - hallway, art gallery, and vending machines. 

\begin{table}[h!]
\centering
\begin{center}
 \begin{tabular}{||c c c c||} 
 \hline
 Class & Train & Validation & Test \\ [0.5ex] 
 \hline\hline
 Art Gallery & 1080 & 360 & 100 \\ [0.5ex]
 \hline
 Hallway & 804 & 268 & 100 \\ [0.5ex]
 \hline
 Other & 793 & 265 & 100 \\ [0.5ex]
 \hline
 Vending Machine & 602 & 201 & 100 \\ [0.5ex]
 \hline\hline
 Total & 3279 & 1094 & 400 \\ [0.5ex]
 \hline
\end{tabular}
\end{center}
\caption{Amount of data collected for training and testing.}
\label{table:dataset}
\end{table}

Other social contexts do not depend on the environmental features, but depend on the non-verbal spatial communication among people -- for example, social contexts like \textit{waiting in a queue} and \textit{O-formations when joining a group}. To account for such non-verbal spatial communication, we collected both simulation and real-world data of people standing in a queue and O-formations using a laser scanner. We collected approximately 170 samples of each context (173 queue and 168 O-formation). A total of 341 samples, split into 80\% training and 20\% test data, are collected both from simulations and real-world interactions.

For the real-world samples, the \textit{leg\_tracker} package~\cite{leigh2015person} detected the positions of people that were later used to calculate circularity and linearity features to train a Support Vector Machines (SVM) model to distinguish between standing in a queue and group formations. 

\subsection{Context Model}
\begin{figure}[h]
    \centering
    \includegraphics[width=0.9\columnwidth]{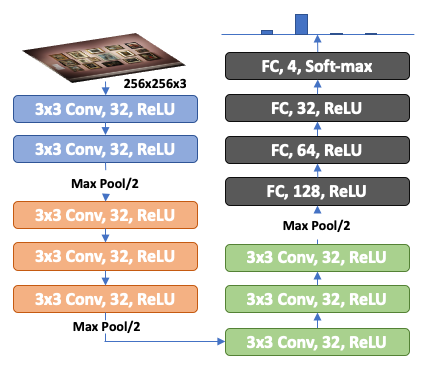}
    \caption{USAN Context Classifier neural network architecture with 8 convolution layers, 3 max-pooling layers and 4 fully connected layers.}
    \label{fig:arch}
\end{figure}

USAN can utilize context information to properly select the objectives specific to the sensed context for a low-level planner~\cite{forer2018socially} to work with. Our approach to a context classifier is a mix of classical machine learning and neural networks. For contexts that include environmental features like hallways, we used images with a CNN architecture that resembles VGGnet~\cite{bojarski2016end} but with a shallow depth. For contexts that depend on non-verbal spatial communication like waiting in a queue, we used laser scanner data with a linear SVM. The CNN takes a 3-channel color image as input and outputs a probability that the image belongs to one of the four classes, as shown in Figure~\ref{fig:arch}. The proposed CNN model consists of 8 convolution layers each with 32 filters, a kernel size of 3, a stride of 1x1, same padding, and ReLU activation. There are three max-pooling layers with a pool size of 2x2 to downsample between layers 2-3, 5-6, 8-9 as shown in Figure~\ref{fig:arch}. The network also includes dropout regularization with every max-pooling layer and between layers 9 and 10 (between first two fully connected layers). All the fully connected layers use ReLU activation expect for the last layer which uses soft-max activation to make the predictions. 

%\begin{figure}[t]
%    \centering
%    \includegraphics[width=\columnwidth]{cnn_roll.png}
%    \caption{Plot showing the performance of the CNN vs. CNN with a rolling mean in comparison with ground truth (annotated by a human) when transitioning between Art Gallery and Hallway contexts.}
%    \label{fig:cnn_roll}
%\end{figure}

When applied to video classification task (continuous frames), the CNN model produced flickering predictions of the scene, a common problem in video classification. We used a rolling average method on the prediction probabilities to get a smooth prediction result of the scene. 
%The rolling average of prediction over N frames. The flickering problem and results of the rolling average method are shown in Figure~\ref{fig:cnn_roll}. The human-annotated data in the figure serves the purpose of ground truth that was annotated by one of the authors based on when the video transitioned from one context to the other. The blue line in Figure~\ref{fig:cnn_roll} represents plain CNN classification results which has a lot of flickering predictions. The orange line represents smoothed out results using rolling average method.

As discussed earlier, there are some social contexts, such as \textit{group formations} and \textit{waiting in queue} which are difficult to be studied by 2-dimensional on-board cameras. However, laser data can be used to understand spatial communication~\cite{banisetty2016socially, sebastian2017socially-aware}, so we used laser scan data to detect and track people in a scene~\cite{leigh2015person}. The positions of the tracked people were used to calculate the following features which were later used in training a linear SVM to distinguish between \textit{waiting in line} and \textit{group formations}:

\textbf{Circularity:} It is used to describe how close a set of points should be to a true circle. The circularity of an irregular polygon formed by a set of points is given by:
\begin{equation}
   C = (4*\pi*area)/perimeter^2
\end{equation}
 Where, area and perimeter of an irregular polygon are:

$area = 1/2\sum{x_{i+1}*y_i - y_{i+1}*x_i}$\\

$perimeter = \sum\sqrt{{(x_{i+1}*y_i)}^2 - {(y_{i+1}*x_i)}^2}$ \\

\textbf{Linearity:} It is the property by which a set of points can be graphically represented as a straight line. The linearity of a set of points is given by:

\begin{equation}
   L = \frac{\sum xy - \frac{\sum x\sum y}{n}}{{\sum x^2}-\frac{({\sum x})^2}{n}}
\end{equation}
Where, n is the number of points/people. 

The range of values for C and L are [0, 1]. People forming a group (circle-like) will have a C value towards 1 and L value towards 0. People forming a line will have a C value towards 0 and L value towards 1. With Circularity and Linearity features, the data is linearly separable, and hence, a linear SVM is one of the simple and ideal models for such data. 

The CNN model using camera input and the SVM model using laser data are two distinct models. Depending on the confidence scores, the cardinal objectives are selected for that particular context. When the detected context is ``Other'', the planner switches to a sub-optimal traditional behavior. 

Scikit-Learn~\cite{scikit-learn} and Keras~\cite{chollet2015keras} with Tensorflow~\cite{45381} backend was used to implement the proposed context classifier (SVM and CNN). The models were built on a computer with an Intel Core i7-8700K CPU @ 3.70GHz x 6 cores, 32 GB of RAM and GeForce GTX 1070 Ti GPU with 8GB memory. The CNN model was trained for 500 epochs with a batch size of 64 on the GPU and took approximately two hours. The model was evaluated for accuracy; the training process included Adam optimizer with categorical cross entropy loss function. The SVM model for spatial data is build on the same hardware with linear support vector classification kernel.

\subsection{Object Detection and Tracking}
%Talk about YOLO-V3 training, dataset, math for pixel to theta, leg tracker.

For detection and tracking of people using a laser scanner, we used a people tracker package~\cite{leigh2015person} by Leigh \textit{et al.} To visually detect and track picture frames (for art gallery interactions), we trained YOLO-v3~\cite{yolov3} on Open Images Dataset. To track picture frames in 3D, we used (x, y) pixel locations in the camera to calculate the depth in the laser scanner data. 

\begin{figure}[t]
    \centering
    \includegraphics[width=0.9\columnwidth]{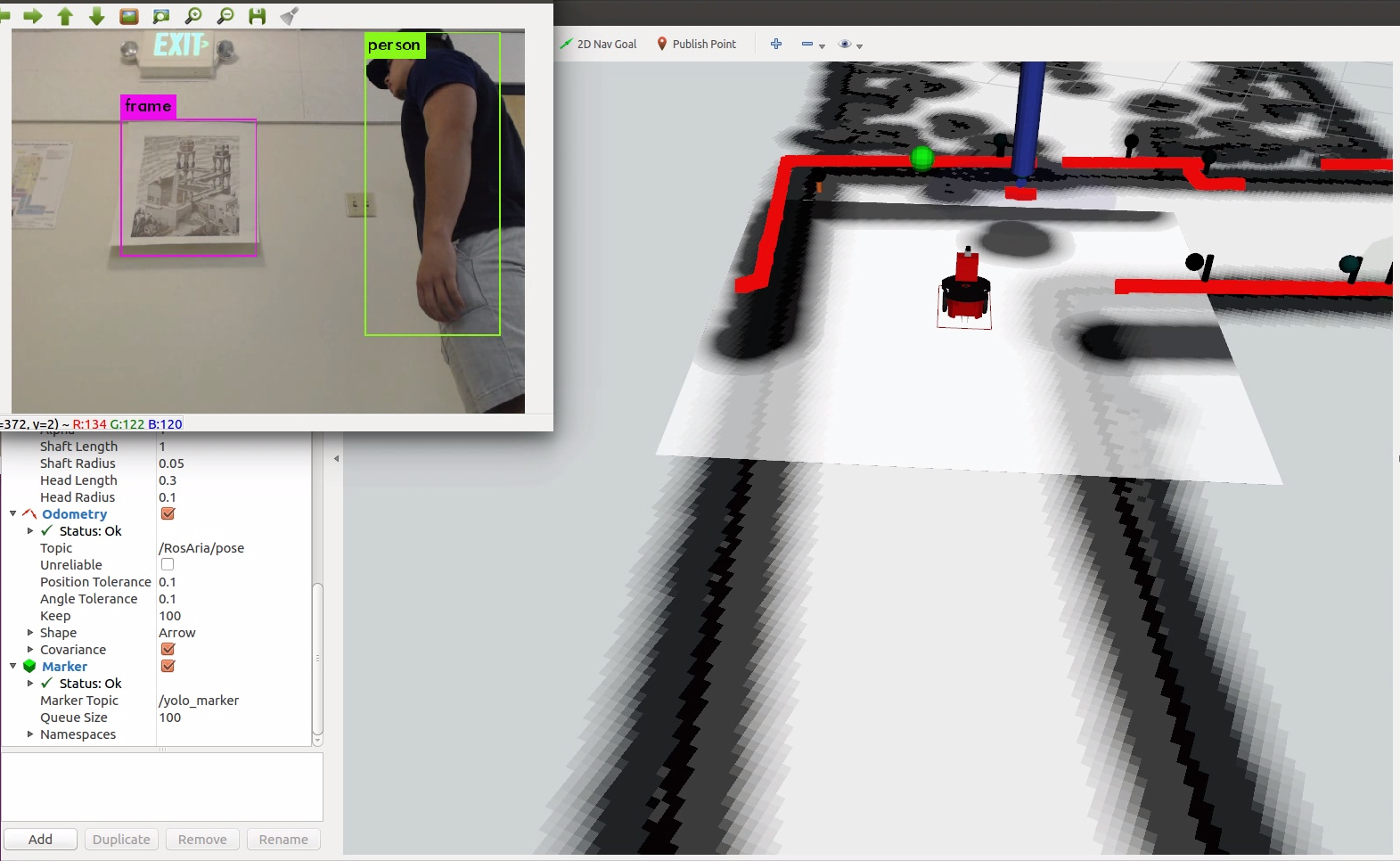}
    \caption{Image showing object detection and tracking using YOLO-v3 and \textit{leg\_tracker} package. The image window (top-left) shows artwork and human detection using YOLO-v3. RVIZ screenshot shows human detection (dark blue cylindrical marker) using \textit{leg\_tracker} package and localization of artwork in laser data (green spherical marker).}
    \label{fig:yolo}
\end{figure}

%Focal length, $f$, of the camera can be calculated using the following equation:

%\begin{equation}
%    f = \frac{x/2}{tan(\frac{1}{2}\phi)}
%\end{equation}
%where, \\
%$x$ = Width of the image in pixels\\
%$\phi$ = Horizontal field of view \\

%Angle made by the centroid of detected picture frame (x,y) pixel value is then used to calculate $\theta$, the angle made by the 3D vectors $a$, $b$ using equation~\ref{eq:theta}.

%\begin{equation}
%\label{eq:theta}
%    \theta = \arccos{\frac{a.b}{|a||b|}}
%\end{equation}
%where, \\
%$\theta$ = angle made by the centroid of the detected object in the camera's field of view\\
%$a$ = 3D vector formed by the center of the image plane and focal length $(0, 0, f)$\\
%$b$ = 3D vector formed by the centroid of the detected object from the optical center and focal length $(d_x, d_y, f)$

%The angle $\theta$ is used to get the depth value from the calibrated laser scan data. The depth value is simply the range value of the laser data at $theta$ angle. Figure~\ref{fig:yolo} shows the visual detection and tracking of picture frames; the green spherical marker in RVIZ shows the 3D location of the tracked picture frame in laser data of the environment. 

\subsection{PaCcET Local Planner}
We use the global trajectory planner, and low-level collision detector~\cite{marder2010office} and make adaptations to the local trajectory planner to incorporate interpersonal distance features using PaCcET. After the context classifier determines the high-level decision of navigational context, the cardinal objectives that matter most are selected. Selected objectives are then utilized by our modified local planner to account for social norms to socially navigate an environment.  

The modified local planner~\cite{forer2018socially, banisetty2019sociallyaware} using PaCcET~\cite{yliniemi2014paccet} can be summarized as follows:

\begin{enumerate}
    \item Discretely sample the robot control space.
    \item Depending on the type of the robot, for each sampled velocity ($V_x$, $V_y$ and $V_{theta}$) perform a forward simulation from the robot's current state for a short duration to see what would happen if the sampled velocities were applied.
    \item Score the trajectories based on metrics.
    \begin{enumerate}
        \item Score each trajectory from the previous step for metrics like distance to obstacles, distance to a goal, etc. Discard all the trajectories that lead to a collision in the environment. 
        \item \label{paccet} For all the valid trajectories, calculate the social objective fitness scores like interpersonal distance and other social features and store all the valid trajectories.
    \end{enumerate}
    \item Perform Pareto Concavity Elimination Transformation (PaCcET) on the stored trajectories to get a PaCcET fitness score and sort the trajectories from lowest to highest PaCcET fitness score.
    \item For each time step, select the trajectory with the highest fitness score.   
\end{enumerate}

In the above working illustration of our low-level planner, step \ref{paccet} is where the social objectives are accounted for while choosing the future valid trajectory points. These social objectives change from context to context and are given by the context classifier module for an autonomously sensed interaction context. 

\section{RESULTS}
\label{sec:results}
\subsection{Perception}
\label{sec:perception}
Our CNN based context classification model was evaluated on validation data, unseen test data and real-world test data. The results are shown in sections \ref{results: validation}, \ref{results: unseen} and \ref{results: real} respectively. The results of the SVM model distinguishing \textit{waiting in a queue} and \textit{O-formations} are presented in section \ref{results: svm}. To validate our context classifier, we used the following metrics:
\begin{itemize}
    \item Confusion matrix, defined as a matrix with elements $C_{ij}$ representing the percentage of observations known to be in class $i$ but predicted as class $j$. For a good classifier, the main diagonal elements should have the highest percentage. 
    \item Precision, intuitively defined as the ability of a classifier not to label a negative sample as positive. It is the ratio $t_p / (t_p + f_p)$.
    \item Recall, intuitively defined as the ability of a classifier to find all the positive samples. It is the ratio of $t_p / (t_p + f_n)$.
    \item F-1 score, can be interpreted as a weighted harmonic mean of the precision and recall. Where, 1 being best and 0 being worst.
\end{itemize}

where, $t_p$ is the number of true positives, $f_p$ the number of false positives and $f_n$ the number of false negatives.

\subsubsection{Validation Set}
\label{results: validation}

\begin{figure}[ht]
    \centering
    \includegraphics[width=0.95\columnwidth]{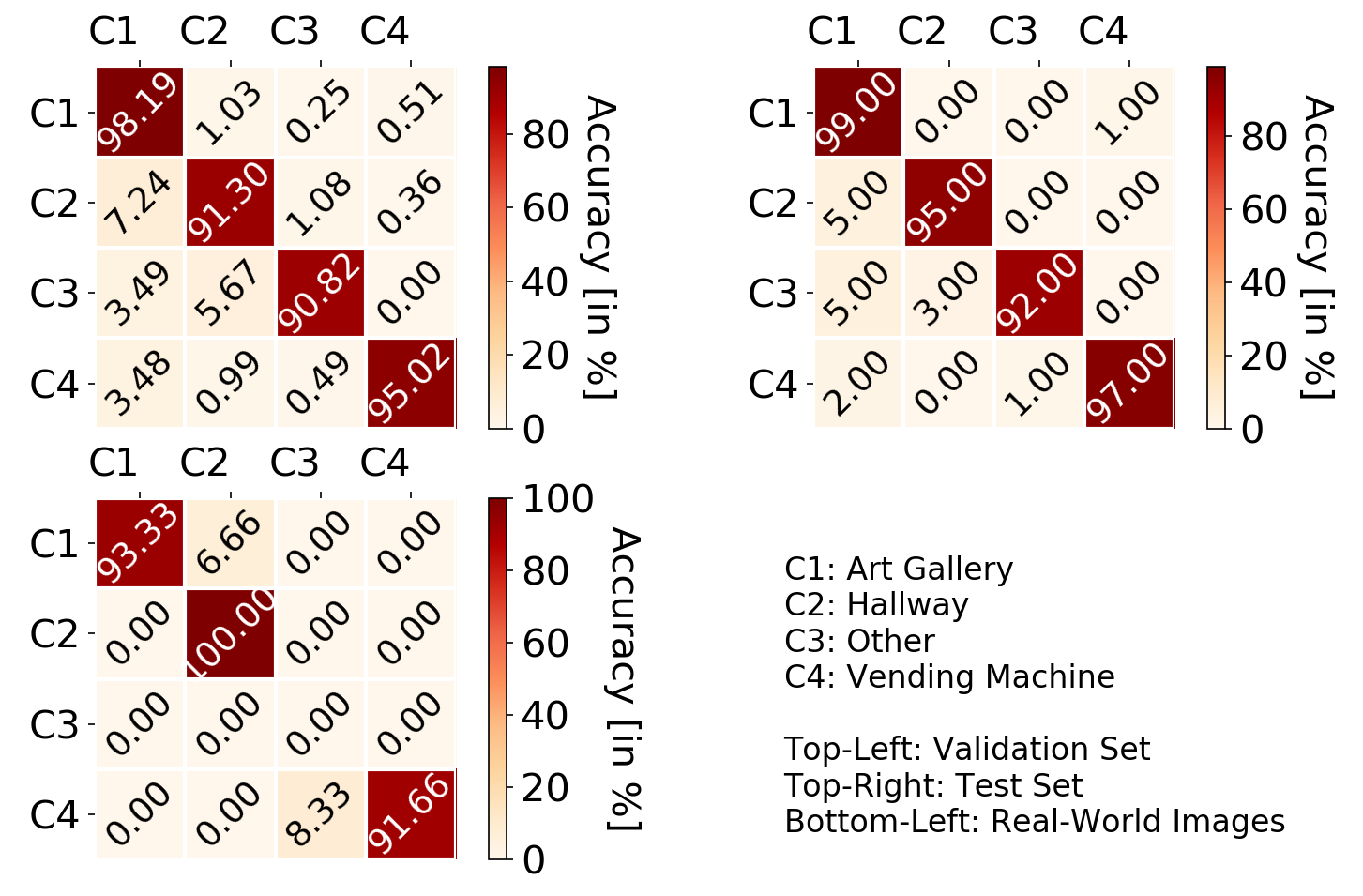}
    \caption{Confusion matrix of validation set, test set and real-world images, showing accuracy (in percentage) for all four context.}
    \label{fig:conf_matrix}
\end{figure}

The model was trained on the training set and validated on the validation set over 500 epochs. Our model achieved a 96.44\% training accuracy and 94.33\% accuracy on validation data.

%The model was trained on the training set and validated on the validation set over 500 epochs, the results of loss and accuracy are shown in Figure~\ref{fig:curves}. Figure~\ref{fig:loss_curves} shows a plot of training (red) and validation loss (blue) for all the epochs. The loss curves show that the model converges over time. Figure~\ref{fig:accuracy_curves} shows a plot of training and validation accuracy for all the epochs. Our model achieved a 96.44\% training accuracy and 94.33\% accuracy on validation data. 

% \begin{figure*}[ht]
%  \centering
% \begin{subfigure}{0.90\columnwidth}
% \includegraphics[width=0.95\columnwidth, height=5cm]{loss_curves.png} 
% \caption{Loss Curves: validation loss (blue) and training loss (red) vs Epochs}
% \label{fig:loss_curves}
% \end{subfigure}
% \hspace{4mm}
% \begin{subfigure}{0.90\columnwidth}
% \includegraphics[width=0.95\columnwidth, height=5cm]{accuracy_curves.png}
% \caption{Accuracy curves: validation accuracy (blue) and training accuracy (red) vs Epochs}
% \label{fig:accuracy_curves}
% \end{subfigure}
% \caption{Training and validation curves: categorical cross entropy loss and accuracy curves}
% \label{fig:curves}
% \end{figure*}

The confusion matrix of the validation set, shown in Figure~\ref{fig:conf_matrix} shows that the model was able to learn to distinguish between an art gallery, a hallway, vending machine, and other contexts with accuracy of 98.19\%, 91.30\%, 95.02\%, and 90.82\% respectively. Table~\ref{table:metrics} shows performance on the validation set.

\begin{table}[h!]
\centering
\begin{center}
 \begin{tabular}{||c c c c||} 
 \hline
 \multicolumn{4}{||c||}{Validation set / Test set / Real-world data}\\ [0.5ex]
 \hline
 Class & Precision & Recall & F1-Score \\ [0.5ex] 
 \hline\hline
 C1 & 0.92 / 0.89 / 1.0 & 0.98 / 0.99 / 0.93 & 0.95 / 0.94 / 0.97 \\ [0.5ex]
 \hline
 C2 & 0.93 / 0.97 / 0.97 & 0.91 / 0.95 / 1.0 & 0.92 / 0.96 / 0.99 \\ [0.5ex]
 \hline
 C3 & 0.98 / 0.99 / 0.00 & 0.91 / 0.92 / 0.00 & 0.94 / 0.95 / 0.00 \\ [0.5ex]
 \hline
 C4 & 0.98 / 0.99 / 1.0 & 0.95 / 0.97 / 0.92 & 0.97 / 0.98 / 0.96 \\ [0.5ex]
 \hline\hline
 \multicolumn{4}{||c||}{C1: Art Gallery, C2: Hallway, C3: Other, C4: Vending Machine}\\ [0.5ex]
 \hline
\end{tabular}
\end{center}
\caption{Performance of the CNN based context classifier.}
\label{table:metrics}
\end{table}

\subsubsection{Unseen Test Set}
\label{results: unseen}

The confusion matrix of the unseen test set (Images from the internet that we kept aside) is shown in Figure~\ref{fig:conf_matrix} shows that the model was able to generalize to unseen data and was able to distinguish between an art gallery, a hallway, vending machine, and other contexts with accuracy of 99.00\%, 95.00\%, 97.00\%, and 92.00\% respectively. Table~\ref{table:metrics} shows performance on the unseen test set. 

\subsubsection{Real-World Data}
\label{results: real}
To see if the model generalizes to real-world images that it has not seen, we collected 15 art gallery, 33 hallway, and 12 vending machines, a total of 60 images on campus. The ``other" category is only a place-holder for any other context apart from the learned hallway, art gallery, and vending machine, so we omitted it from this test set. When in an unknown context, the planner can select default, but likely sub-optimal, objectives that will reward safe movement from one place to another. As seen in Figure~\ref{fig:conf_matrix}, the model performed well on real-world images as well. The accuracy for an art gallery, hallway, and vending machine categories are 93.33\%, 100.0\%, and 91.66\%, respectively. The performance on real-world data is presented in Table~\ref{table:metrics}.

\subsubsection{Group and Queue Formations}
\label{results: svm}

\begin{figure}[ht]
    \centering
    \includegraphics[width=0.95\columnwidth]{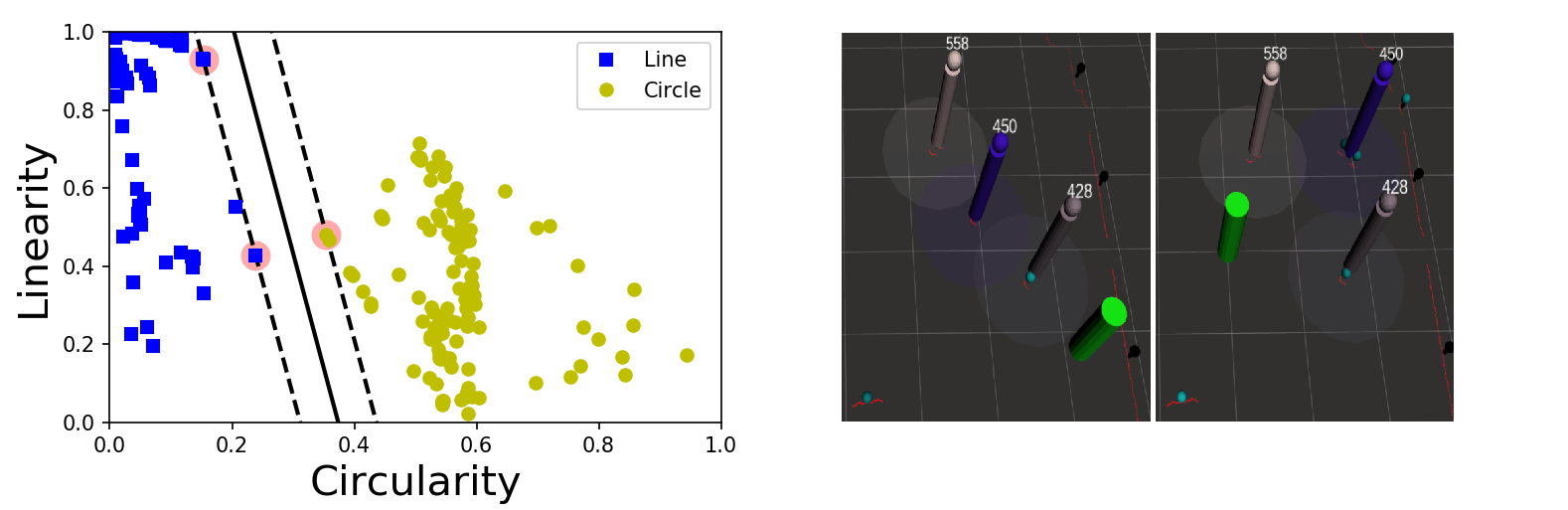}
    \caption{\textbf{Left:} trained SVM classifier, \textbf{Right:} Social goal determined by the robot in \textit{waiting in queue} and \textit{O-formation} contexts.\vspace{-4mm}}
    \label{fig:svm}
\end{figure}

We trained a linear SVM on location data collected from a laser scanner to classify if a group of people as \textit{waiting in a queue} or forming a \textit{O-formation}. We selected features like circularity, linearity, and the radius of the best-fit circle (with standardization). Later, we trained the SVM omitting the radius feature as circularity and linearity are sufficient to differentiate between the two classes, as shown in Figure~\ref{fig:svm} (left). The trained SVM achieved 100\% accuracy on both training and test data.
%, training set results with three-fold cross-validation, and the test set is shown in bottom-left and bottom-right of Figure~\ref{fig:svm} respectively.
Precision, recall, and f1-scores are all 1.00 for both training and test sets. Figure~\ref{fig:svm} (right) shows an rviz screenshot of the computed social goal (green marker) determined by the robot in \textit{waiting in queue} and \textit{O-formation}. 

\subsection{Cardinal Objective Selection}
\label{sec:obj}
\begin{figure}[h]
    \centering
    \includegraphics[width=0.95\columnwidth]{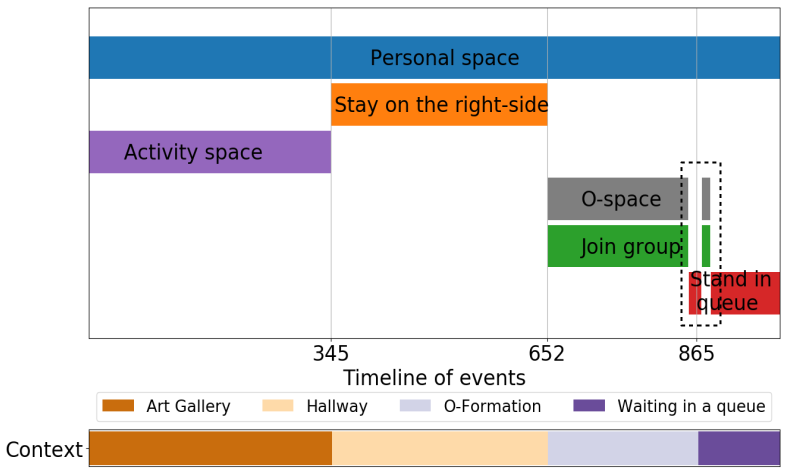}
    \caption{Timeline showing the social objectives selected by the robot when teleoperated in an environment with hallways, artwork, people in O-formations, and people waiting in queue contexts.}
    \label{fig:activation}
\end{figure}

% \blindtext
We teleoperated the robot in an environment with \textit{hallways}, \textit{artwork}, \textit{people in O-formations}, and \textit{people waiting in queues} to test if the models can select objectives related to detected context. The results of the robot deciding on the objectives for an autonomously sensed context are shown in Figure~\ref{fig:activation}, the transitions from one context to the other are shown using the vertical grid lines. Figure~\ref{fig:activation} shows that the robot is considering personal space and activity space in an \textit{art gallery} situation. In a \textit{hallway} situation, the robot accounts for personal space and staying on the right-side objectives. Similarly, in a \textit{group (0-formation)} scenario, the robot considers the personal space of all the people, the O-space of the group, and the social goal of joining the group. In \textit{waiting in a queue} context, the robot considers joining the end of the line along with the personal space of the people forming the line. It is also important to note that reaching the goal, and collision avoidance are other objectives of our PaCcET local planner.

The black box with the dotted line in Figure~\ref{fig:activation} shows the ambiguity of classification during the transition of the same group of people from \textit{O-formation} to a \textit{line formation}. This ambiguity is due to the quick change in the group dynamics, but misclassification for a fraction of a second should not affect the overall social performance of the planner.

\subsection{Socially-Aware Navigation}
In Sections~\ref{sec:perception} and \ref{sec:obj}, we discussed the results of perception pipeline: performance of the CNN based visual classification, SVM based group scenario classification using laser data, by teleoperating the robot in an environment, we showed that our method was able to detect the context accurately and thereby was able to select the cardinal objectives for that particular context. 

\begin{figure}[h]
    \centering
    \includegraphics[width=0.95\columnwidth]{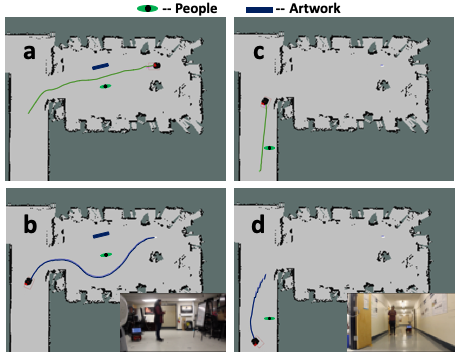}
    \caption{Sub figures \textbf{a}, \textbf{c} shows a non-social path a robot with traditional planner would take in an \textit{art gallery} and \textit{hallway} contexts respectively. Sub figures \textbf{b}, \textbf{d} shows the social path our SAN planner executed.}
    \label{fig:art-hallway}
\end{figure}

Figure~\ref{fig:art-hallway} shows the robot's interaction in an \textit{art gallery} followed by a \textit{hallway} context. In the \textit{art gallery} context, the robot encountered one spectator viewing the art. When switching to hallway context, the robot encountered a person in a narrow hallway. The green trajectory in figure~\ref{fig:art-hallway} \textbf{a}, \textbf{c} represents the shortest global trajectory that a traditional local planner would closely follow. In figure~\ref{fig:art-hallway} \textbf{a}, the trajectory violates the social rule of traversing in the activity zone (space between the artwork and the spectator). In figure~\ref{fig:art-hallway} \textbf{c}, the trajectory violates the personal space around the human in a hallway. On the other hand, in figure~\ref{fig:art-hallway} \textbf{b}, our social planner steered the robot away from the activity space, thereby executing a socially appropriate trajectory in an art gallery. Similarly, in figure~\ref{fig:art-hallway} \textbf{d}, our social planner steered the robot in such a way that it does not violate a person's personal space. 

\begin{figure}[h]
    \centering
    \includegraphics[width=0.95\columnwidth]{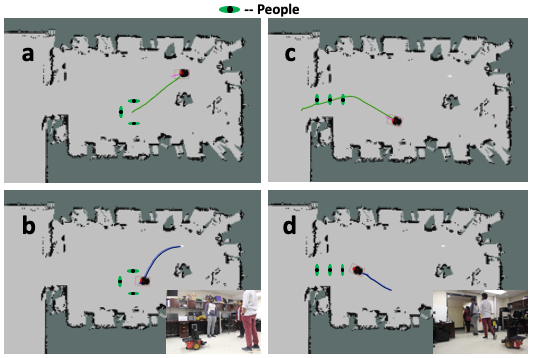}
    \caption{Sub figures \textbf{a}, \textbf{c} shows a non-social path a robot with traditional planner would take in an \textit{O-formation} and \textit{waiting in a queue} contexts respectively. Sub figures \textbf{b}, \textbf{d} shows the social path our SAN planner executed.}
    \label{fig:circle-line}
\end{figure}

Figure~\ref{fig:circle-line} shows the robot's interaction in an \textit{O-formation} situation followed by a \textit{waiting in a queue} context. In both these contexts, the robot interacted with three humans. The green trajectory in figure~\ref{fig:circle-line} \textbf{a}, \textbf{c} represents the shortest global trajectory that a traditional local planner would closely follow. In figure~\ref{fig:circle-line} \textbf{a}, the trajectory planner steered the robot to the center of the group, placing it in an inappropriate location to meet with the group. In figure~\ref{fig:circle-line} \textbf{c}, the generated trajectory forces the robot to cut the line which is socially inappropriate. On the other hand, in figure~\ref{fig:circle-line} \textbf{b}, our social planner steered the robot to an appropriate location on the circle formed by the group (social goal). Similarly, in figure~\ref{fig:circle-line} \textbf{d}, our social planner steered the robot to the end of the line formed by the people (social goal). The social goal calculation in \textit{O-formation} and \textit{waiting in a queue} context is determined by geometric reasoning~\cite{banisetty2019sociallyaware}. 

\section{Discussion and Future Work}
\label{sec:discussion}
%talk about other category and confusion between art gallery and hallway.
%Why did we pick these contexts. 
%talk about traditional ml work for interaction.

%Future work, talk about intent recognition and it role in USAN architecture.  

%\blindtext[3]

Our prior work~\cite{forer2018socially} proposed a non-linear multi-objective optimization based PaCcET local planner using two objectives that was able to execute socially-aware behavior in a hallway setting. We then extended it to include more than two objectives to show that our PaCcET local planner can scale and extend to complex social situations like avoiding activity zones, joining a group, and waiting in a line scenarios~\cite{banisetty2019sociallyaware}. In this paper, we concentrate on the PaCcET-enabled local planner in conjunction with a hybrid context classification method using CNN and SVM to demonstrate that architecture shown in Figure~\ref{fig:usan} can be used to exhibit socially-aware navigation behaviors in multiple social contexts.

Real-world long-term deployment of service robots require a unified socially-aware navigation method that can exhibit social navigation behavior in every social situation it might encounter in a dense human environment. Our proposed work is novel yet has certain limitations/improvements that can push USAN methods in real-world deployment. Possible improvements and future work includes the following:
\begin{enumerate}
    \item The trained CNN classifier works well for the trained contexts but a better solution would be a combination of learning and reasoning. For example, the model learns what objects constitute a context, later when encountered a situation, it should reason about the correct context against a knowledge base from prior experience.
    
    Our ongoing efforts include building a broader knowledge base using MIT Indoor Scenes dataset~\cite{indoorscenes}. Future work will augment our system to autonomously build a knowledge graph by learning the relationships between contexts and objects within the context~\cite{salek2021towards}.
    
    \item The cardinal objectives are hand-picked for each trained context. Possible improvement would be to learn these objectives from human-human interactions without being explicitly told.
    \item When closely observed, human-human navigational interaction benefits from intent communication and intent recognition. An intent module that can both infer and communicate navigational intentions would make our proposed method predictive system as opposed to a reactive system. 
\end{enumerate}

Real-world deployment of social robots that can socially navigate in a human dense human-robot environment may be far off. But it is clearly evident that social behavior in one context is not sufficient for long-term acceptance of service robots in public place. With this work, we demonstrate how differing navigation behavior is appropriate given different social and environmental contexts and that visual and laser range information can be used to autonomously sense the context.

\section{Conclusion}

It is unlikely that social behavior for a single context is sufficient for long-term acceptance of service robots in public places. As robots are increasingly present in human environments, these robots need to account for social norms in various navigational contexts. There is a need for a unified architecture that can autonomously sense the ongoing navigational interaction and execute a trajectory that is socially appropriate for that particular interaction context. We presented a novel approach to a unified socially-aware navigation, discussed various subsystems, and implemented it on a robot. In this paper, we showed that a context classifier along with a low-level planner utilizing PaCcET could be used to generate socially optimal trajectories for an autonomously sensed social context. The perception system has generalized to new data and had performed well in recognizing the contexts in real human environments. The navigation results show that the robot was able to account for the social norms while performing navigational actions in various social contexts such as hallway interactions, art gallery situations, O-formations when joining a group, and waiting in queue situations.

%%%%%%% SBB READ TILL HERE Feb 26 %%%%%%%%

%                                   % the textheight of the last page by a suitable amount.
%                                   % This command does not take effect until the next page
%                                   % so it should come on the page before the last. Make
%                                   % sure that you do not shorten the textheight too much.

%%%%%%%%%%%%%%%%%%%%%%%%%%%%%%%%%%%%%%%%%%%%%%%%%%%%%%%%%%%%%%%%%%%%%%%%%%%%%%%%

%%%%%%%%%%%%%%%%%%%%%%%%%%%%%%%%%%%%%%%%%%%%%%%%%%%%%%%%%%%%%%%%%%%%%%%%%%%%%%%%



\section*{Acknowledgement}
The authors would like to acknowledge the financial support of this work by the National Science Foundation (NSF, \#IIS-1719027, IIS-1757929), Nevada NASA EPSCoR (\#NNX15AI02H). We would like acknowledge the help of Athira Pilla, Ashish Kasar, Anuraag Gupta, and Mounica Santhapur.

\bibliographystyle{ieeetr}
\bibliography{bibo}

\end{document}